\documentclass[conference]{IEEEtran}
\IEEEoverridecommandlockouts

\usepackage{cite}
\usepackage{amsmath,amssymb,amsfonts}
\usepackage{algorithmic}
\usepackage{graphicx}
\usepackage{textcomp}
\usepackage{xcolor}
\usepackage{booktabs}
\usepackage{multirow}
\usepackage{makecell}
\usepackage{amssymb} 
\usepackage{pifont}
\usepackage{xcolor}
\usepackage{booktabs}
\usepackage{siunitx}
\usepackage{threeparttable}
\def\BibTeX{{\rm B\kern-.05em{\sc i\kern-.025em b}\kern-.08em
    T\kern-.1667em\lower.7ex\hbox{E}\kern-.125emX}}
\begin{document}

\title{BioBridge: Bridging Proteins and Language for Enhanced Biological Reasoning with LLMs}

\author{
    \IEEEauthorblockN{
        Yujia Wang\IEEEauthorrefmark{2},
        Jihong Guan\IEEEauthorrefmark{2}\IEEEauthorrefmark{1},
        Wengen Li\IEEEauthorrefmark{2},
        Shuigeng Zhou\IEEEauthorrefmark{3},
        Xuhong Wang\IEEEauthorrefmark{4}\IEEEauthorrefmark{1}
    }
    \IEEEauthorblockA{\IEEEauthorrefmark{2}School of Computer Science and Technology, Tongji University, Shanghai, China\\
    \{2432094, jhguan, lwengen\}@tongji.edu.cn}
    \IEEEauthorblockA{\IEEEauthorrefmark{3}College of Computer Science and Artificial Intelligence, Fudan University, Shanghai, China\\
    sgzhou@fudan.edu.cn}
    \IEEEauthorblockA{\IEEEauthorrefmark{4}Shanghai AI Laboratory, Shanghai, China\\
    wangxuhong@pjlab.org.cn}
    \thanks{\IEEEauthorrefmark{1} Corresponding authors.}
    \thanks{This work was supported in part by the National Nature Science Foundation of China (No.62372326,No.62172300),and the Shanghai Municipal Education Commission Artificial Intelligence-Driven Research Paradigm Reform and Discipline Leapfrogging Empowerment Program (No.2024RGYB001),the China Postdoctoral Science Foundation (No.2025M771493).}
}

\maketitle

\begin{abstract}
Existing Protein Language Models (PLMs) often suffer from limited adaptability to multiple tasks and exhibit poor generalization across diverse biological contexts. In contrast, general-purpose Large Language Models (LLMs) lack the capability to interpret protein sequences and fall short in domain-specific knowledge, limiting their capacity for effective biosemantic reasoning. To combine the advantages of both, we propose BioBridge, a domain-adaptive continual pretraining framework for protein understanding. This framework employs Domain-Incremental Continual Pre-training (DICP) to infuse protein domain knowledge and general reasoning corpus into a LLM simultaneously, effectively mitigating catastrophic forgetting. Cross-modal alignment is achieved via a PLM-Projector-LLM pipeline, which maps protein sequence embeddings into the semantic space of the language model. Ultimately, an end-to-end optimization is adopted to uniformly support various tasks, including protein property prediction and knowledge question-answering. Our proposed BioBridge demonstrates performance comparable to that of mainstream PLMs on multiple protein benchmarks, such as EC and BindingDB. It also achieves results on par with LLMs on general understanding tasks like MMLU and RACE. This showcases its innovative advantage of combining domain-specific adaptability with general-purpose language competency. 
\end{abstract}


\section{Introduction}

\begin{figure}[t]
\centerline{\includegraphics[width=0.5\textwidth]{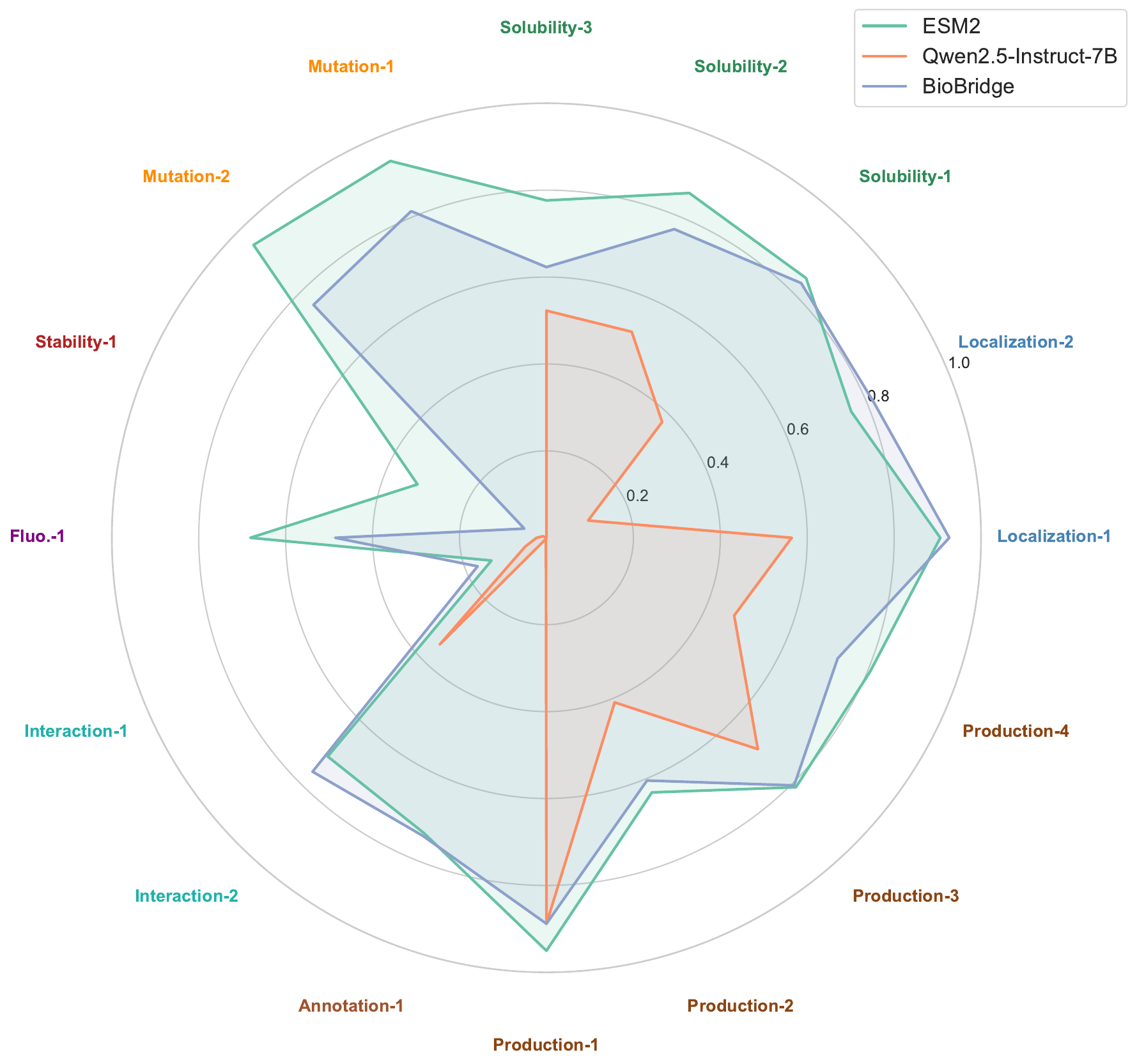}}
\caption{BioBridge Unlocks PLM-Level Performance for Qwen in Protein Research.}
\label{fig: radar}
\end{figure}
As central players in life's activities, proteins constitute a core class of biotherapeutic drugs. However, a significant bottleneck impedes target identification and drug discovery: less than 0.3\% of protein sequences in the UniProt database have reliable experimental functional annotations. Therefore, efficient prediction of protein properties (function, stability, mutational effects, structure) is critical, and building a unified multi-task protein understanding system would advance disease research, drug development, and synthetic biology.

In recent years, Protein Language Models (PLMs)—rooted in Transformer architectures and trained via self-supervised learning on massive unlabeled protein sequences—have achieved breakthroughs in structure prediction, functional annotation, and mutational modeling \cite{b1}. Methods like Tranception (autoregressive sequence reconstruction) and ESM \cite{b3} (masked language modeling, MLM) capture residue relationships and structural awareness, offering a new paradigm for decoding protein "language" \cite{b3}. However, current PLMs are confined to isolated downstream tasks, lacking generalization and deep modeling of sequence-structure-function causal relationships, which limits scientific interpretability.

Thus, advancing PLMs from "single-task experts" to "multi-task intelligent agents" is crucial, and integrating Large Language Models (LLMs)—with strong reasoning and cross-scenario generalization—is essential. While models like ProtT3\cite{b11}, Prot2Text\cite{b16}, and Prot2Chat \cite{b31} have explored protein-natural language fusion, they rely on small LLMs (e.g., Galactica-1.3B, GPT2-1.5B) and task-specific training, resulting in limited generality, narrow task coverage, and rigid input/output constraints.


\begin{figure}[!htbp]
\centerline{\includegraphics[width=0.5\textwidth]{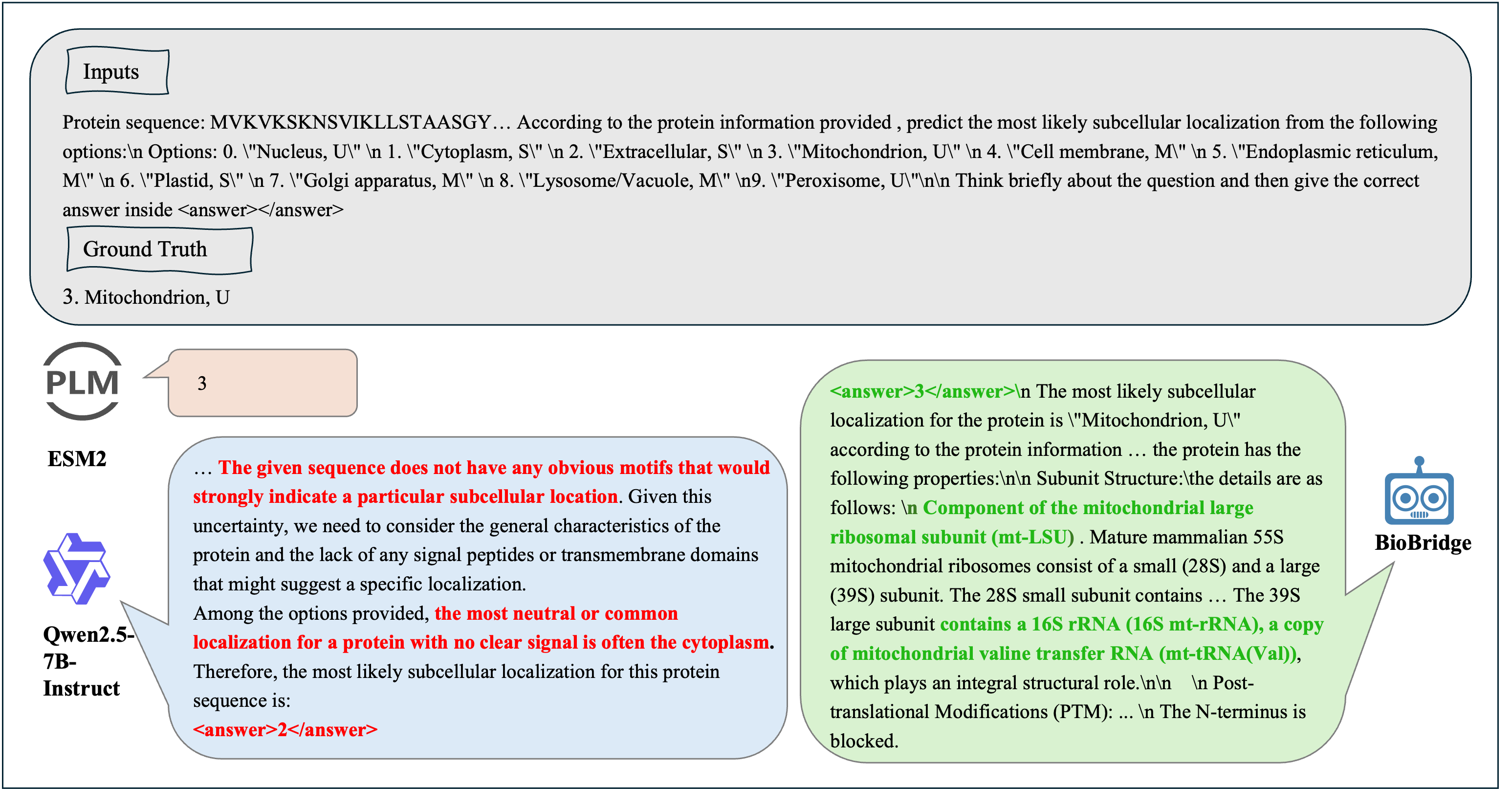}}
\caption{\textbf{Case Study.}Red fonts represent incorrect answers, and green fonts represent correct key answers.}
\label{fig: case}
\end{figure}

Two core challenges hinder effective PLM-LLM fusion: (1) \textit{Biological Knowledge Barrier and Catastrophic Forgetting}: Mainstream LLMs lack systematic biological knowledge, and while continual pre-training (CPT)—which supplements domain knowledge without full retraining \cite{b4}—can mitigate this, it often causes catastrophic forgetting of original linguistic capabilities \cite{b10}. Existing CPT methods (e.g., distillation for corpus adaptation \cite{b6}, parameter-efficient fine-tuning \cite{b9}, domain-aligned pre-training \cite{b7}) struggle to balance old and new knowledge retention, and often require post-CPT instruction recovery \cite{b12}. (2) \textit{Modality Difference Modeling}: Protein sequence grammar differs fundamentally from natural language, and function depends on long-range interactions. While contrastive learning (e.g., CLIP-inspired protein-text pairing \cite{b13,b14}, structural/functional modality integration \cite{b16,b18}, Transformer-based contextual alignment \cite{b19,b20}) aids cross-modal representation alignment, it remains surface-level and fails to enable LLMs to reason using intrinsic biological knowledge.

In this work, we propose BioBridge, a multi-task protein understanding framework that addresses current protein-domain LLM limitations—specifically inadequate knowledge transfer and modality misalignment—and enhances task generalization. BioBridge outperforms the general LLM Qwen2.5-7B-Instruct, approaches the performance of the dedicated PLM ESM2 (Figure \ref{fig: radar}), and excels in subcellular localization prediction by delivering accurate results alongside scientific explanations(Figure \ref{fig: case}). Its key innovations are as follows: (1) Domain-Incremental Continual Pre-training (DICP): This strategy injects biomedical knowledge into general LLMs via specialized biomedical corpora, and ablation studies confirm that it enables robust domain adaptation without compromising the models’ general language understanding; (2) PLM-Projector module: It uses ESM2 as a protein encoder to extract sequence representations, and integrates a cross-modal Projector that maps protein embeddings to the LLM’s text semantic space—this enhances protein-text alignment and strengthens the LLM’s ability to understand protein data; (3) Modality alignment for integration: The two aforementioned stages are integrated through modality alignment, which establishes cross-modal connections between protein sequences and natural language tokens via end-to-end fine-tuning, ultimately enabling unified multi-modal learning.

\begin{figure*}[htbp]
\centerline{\includegraphics[width=1\textwidth]{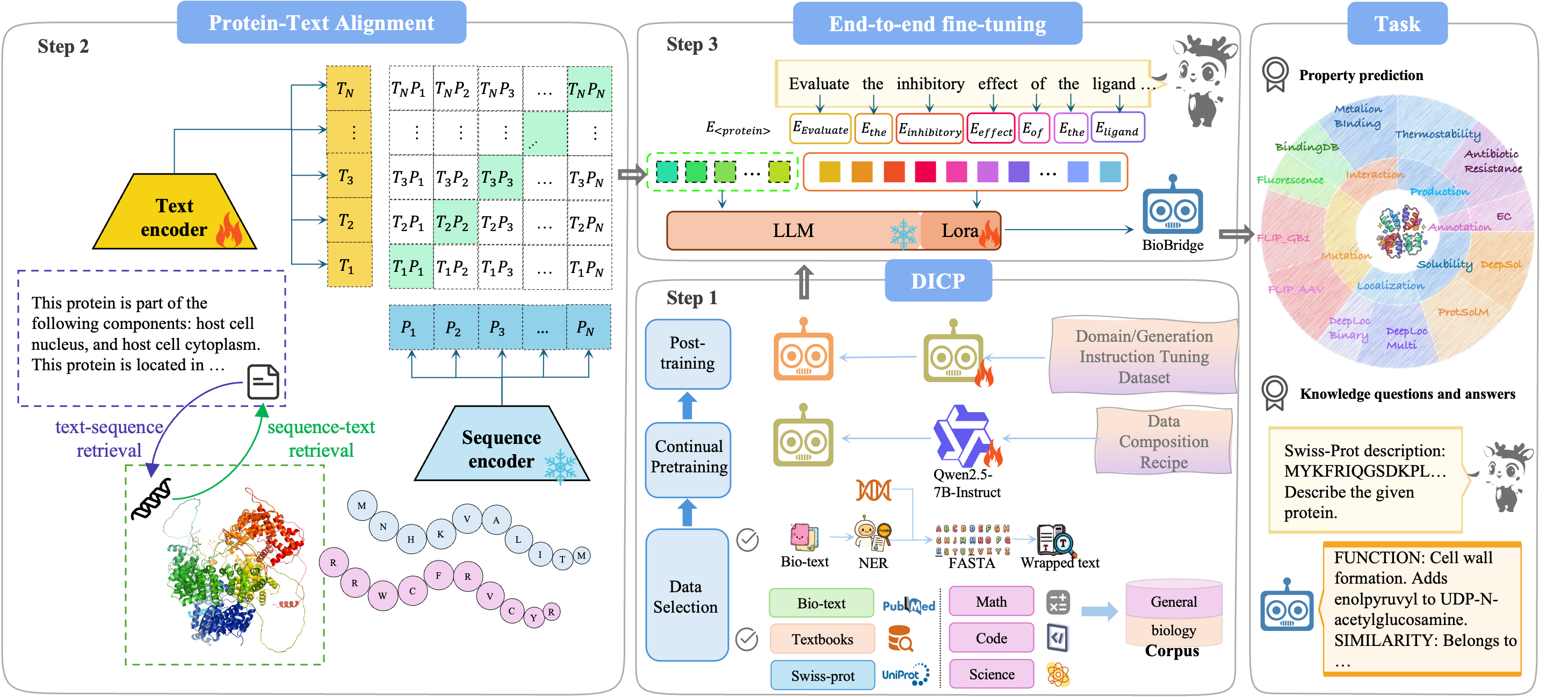}}
\caption{\textbf{illustration of our model}: Domain-Incremental Continual Pre-training (DICP) adapts a general language model to biomedical data via continual pretraining on unlabeled domain-specific corpora. PLM-Projector maps protein embeddings into the language model’s space for protein-text alignment. End-to-End Fine-tuning connects protein and text tokens through joint optimization. Versatile Applications include alignment, classification, and generation tasks.
}
\label{fig: framework}
\end{figure*}

\section{Materials and Methods}

\subsection{problem formulation}
The core challenge is aligning the representations of natural language and protein language.We use a dataset $\mathcal{D}=\{(\mathbf{p}_i,\mathbf{t}_i)\}_{i=1}^N$ comprising protein sequences ($\mathbf{p}_i$) and their text descriptions ($\mathbf{t}_i$).

Our objective is cross-modal alignment in a shared semantic space $\mathcal{Z}$.We introduce a protein token extractor $f_p$ and a text encoder $f_t$, where the protein output is mapped to $\mathcal{Z}$ via a projection layer.
Alignment is achieved using contrastive loss to minimize the distance between positive pairs $(\mathbf{z}_i^p, \mathbf{z}_i^t)$.

In subsequent multi-modal instruction fine-tuning, protein tasks are embedded into natural language prompts:
\begin{equation}
\mathbf{x}_{i}=[\mathbf{t}_{i,1},\ldots,\mathbf{t}_{i,k},\mathrm{<protein>,~}\mathbf{z}_{i}^{p},\mathrm{</protein>}]
\end{equation}

In the evaluation stage (e.g., protein property prediction $\mathcal{T}_{\mathrm{cls}}$), the input maintains the same multi-modal structure. The model generates a classification prediction $\hat{y}_i$ using a softmax classifier on the final output representation.

\subsection{Overall Framework}
As depicted in Figure \ref{fig: framework}, our model development unfolds across three distinct training stages: (1) Domain-Incremental Continual Pre-training (DICP) within the biomedical domain, (2) Protein-Text Cross-Modal Feature Alignment Modeling, and (3) Protein and text tokens are aligned via joint optimization in an end-to-end fine-tuning process.

\subsection{DICP}
The continual pre-training phase is meticulously designed to deepen the model's comprehension of intricate protein structures and broader biological knowledge. To achieve this, our training data amalgamates diverse biological information from four distinct sources, each contributing to a comprehensive multi-modal understanding:
(1) \textbf{Advanced Biology Textbook Texts}: This compilation includes core knowledge from pivotal fields such as molecular biology, biochemistry, and genetics, providing foundational scientific context.
(2) \textbf{PubMed Central Open Access Articles and PubMed Abstracts}: We incorporate the full texts of English biomedical papers and medical abstracts harvested, offering extensive domain-specific literature.
(3) \textbf{Bio-sequence Injected Text}: Utilizing PubMed articles \cite{b21}, we employ BERN2 \cite{b22} for named entity recognition. This allows us to either replace or append corresponding amino acid sequences, embedding them with the sequence token identifier \texttt{<seq>}. Scientific sentences not augmented with biological sequences are retained as general text to further enrich the corpus.
(4) \textbf{Protein and Description Pairs}: We source high-quality protein FASTA description pairs from 90K Swiss-Prot \cite{b23}, which provide rich annotations across various protein characteristics.

\sisetup{table-number-alignment=center, round-mode=places, round-precision=2}
\begin{table}[t]
\caption{Composition of the protein domain continuous pre-training set}
\centering
\setlength{\tabcolsep}{4mm}
\renewcommand{\arraystretch}{1.2}
\begin{tabular}{@{}l
                S[table-format=4.2]
                S[table-format=3.2]
                @{}}
\toprule
\textbf{Data Source} & \multicolumn{1}{c}{\textbf{Token (M)}} & \multicolumn{1}{c}{\textbf{Ratio (\%)}} \\
\midrule
Biology Textbooks            & 104.19  & 6.05  \\
PubMed Articles              & 553.93  & 32.16 \\
Sequence-Augmented Sentences & 62.11   & 3.60  \\
Protein-Description Pairs    & 343.58  & 19.95 \\
Math                         & 384.72  & 22.34 \\
Code                         & 133.86  & 7.77  \\
Science                      & 140.01  & 8.13  \\
\midrule
\textbf{Total}               & \textbf{1722.40} & \textbf{100.00} \\
\bottomrule
\end{tabular}
\label{tab:pretrain_data}
\end{table}

We unified all data during preprocessing to a maximum token sequence length of 4096. Sequences exceeding this length were truncated at sentence boundaries. Shorter samples were concatenated to fully utilize the context window and avoid meaningless padding.

In the instruction fine-tuning stage, we used the Alpaca-GPT4 dataset \cite{b24}, comprising 52K English instructions, to restore the model's task-following capabilities.


Enhancing domain adaptation through continual pre-training with specialized corpora risks catastrophic forgetting. To counteract this, and inspired by the concept of replay datasets, we fortified our protein-centric corpus with a small amount of high-quality Mixture of Thoughts (MoT) data. This blended dataset, containing 93K mathematical, 83K code, and 173K scientific problems, allows us to maintain general knowledge while specializing in the protein domain.


By incorporating this natural language reasoning data, we enhanced the model's reasoning capabilities, striking a balance between protein domain specialization and general reasoning ability. This improved performance on scientific understanding tasks without compromising core protein representation learning. The final dataset composition is detailed in Table \ref{tab:pretrain_data}.


We utilized Qwen2.5-7B-Instruct as our base model for this training stage. The process was executed in a multi-node configuration, employing 24 NVIDIA A800 (80GB) GPUs across three nodes. Training was performed for one full epoch on the mixed natural language and protein-specific corpus. We adhered to the recommended hyperparameters for Qwen2.5, using a sequence length of 4096 and a learning rate of $1.0e^{-5}$.

\subsection{Protein-Text alignment Training}
In the protein-text alignment phase, we used two large-scale datasets:
(1) \textbf{OntoProtein (422K pairs)}: Contains protein sequences paired with ontology-based textual annotations, covering molecular functions, biological processes, and cellular components.
(2) \textbf{Swiss-Prot (430K pairs)}: Comprises manually curated protein sequences and expert-written descriptions, detailing functional and structural characteristics across diverse organisms.

The objective of this phase is to compress protein sequence information into fixed-length latent vectors. These vectors are designed to serve as input to the LLM and undergo alignment training with natural language text. We achieve this using a structure comprising a protein encoder, a querying transformer, and an alignment contrastive learning objective.

\subsubsection{Protein Encoder}
We adopt ESM2 as our foundational protein model. ESM2 is a Transformer-based language model that leverages attention mechanisms to learn interaction patterns between amino acid pairs within an input sequence. In our work, it serves as the first component of the protein token extractor, denoted as $\text{ProtEnc}(\cdot)$, mapping an amino acid sequence to a context-rich embedded representation sequence of length $L$:
\begin{equation}
\mathbf{h}=\mathrm{ProtEnc}(\mathbf{p})=(\mathbf{h}_{1},\mathbf{h}_{2},\ldots,\mathbf{h}_{L}).
\end{equation}
During the training process, the parameters of the protein encoder are kept frozen to ensure stable sequence representations.

\subsubsection{Querying Transformer}
To obtain a fixed-dimensional protein representation, we introduce a Q-Former \cite{b29} module, denoted as $\text{QFormer}_\phi(\cdot)$. The Q-Former receives $K$ learnable query token vectors $\mathbf{q} = (\mathbf{q}_1, \ldots, \mathbf{q}_K)$ and extracts text-task-relevant features from $\mathbf{h}$ via a cross-attention mechanism, ultimately outputting $K$  latent vectors:
\begin{equation}
\mathbf{z}^p=(\mathbf{z}_{1},\ldots,\mathbf{z}_{K})=\mathrm{QFormer}_{\phi}(\mathbf{h},\mathbf{q})
\end{equation}
This architecture offers greater flexibility compared to traditional \texttt{<cls>} token representations. The adjustable number of latent vectors, $K$, allows for the retention of more structural information, which is beneficial for subsequent alignment learning in natural language tasks.

\subsubsection{Contrastive alignment Objective}
We adopt a cross-modal contrastive learning method \cite{b26} to align the protein representations extracted by the Q-Former with their corresponding text representations. Specifically, let a training batch consist of $B$ protein-text pairs: ${(p_1, t_1), ..., (p_B, t_B)}$. For the i-th protein $p_i$, we denote its representation vector as $\mathbf{z}_i^p \in \mathbb{R}^{N_q \times d}$, where $\mathbf{z}_{ij}^p \in \mathbb{R}^d$ is the representation of the j-th query token. The corresponding text $t_i$'s representation is $\mathbf{z}_i^t \in \mathbb{R}^d$, typically obtained from the output of its [CLS] token.

Our goal is to learn a mapping relationship through the cross-modal alignment network $\mathcal{F}_{\theta}$ such that the protein representation $\mathbf{z}^p$ and its corresponding text representation $\mathbf{z}^t$ are close in a shared semantic space. In other words, the objective is to minimize the distance between positive sample pairs while maximizing the discriminability from negative sample pairs.
\begin{equation}
\begin{gathered}
\mathcal{L}_{\mathrm{p2t}}=\frac{1}{B}\sum_{i=1}^B\log\frac{\exp(\max_k\cos(\mathbf{z}_{ik}^p,\boldsymbol{\mathbf{z}_i^t}/\tau)}{\sum_{j=1}^B\exp(\max_k\cos(\mathbf{z}_{ik}^p,\boldsymbol{\mathbf{z}_j^t}/\tau)}, \\
\mathcal{L}_{\mathrm{t2p}}=\frac{1}{B}\sum_{i\operatorname{=}1}^B\log\frac{\exp(\max_k\cos(\mathbf{z}_{ik}^p,\boldsymbol{\mathbf{z}_i^t}/\tau)}{\sum_{j\operatorname{=}1}^B\exp(\max_k\cos(\mathbf{z}_{jk}^p,\boldsymbol{\mathbf{z}_i^t}/\tau)}, \\
\mathcal{L}_{\mathrm{PTC}}=-\mathcal{L}_{\mathrm{p2t}}-\mathcal{L}_{\mathrm{t2p}}.
\end{gathered}
\end{equation}

For the top few semantically similar protein tokens extracted, a more fine-grained protein-text similarity calculation is employed. The query tokens and text tokens are jointly fed into the same self-attention module of the Q-Former, enabling cross-modal interaction from the early stages of the model. This approach allows the query tokens to not only focus on protein features but also simultaneously capture associated text information during the encoding process. Subsequently, we perform average pooling on the output representations of all query tokens, using this as the joint representation for the protein-text pair, which is then fed into a linear classifier to predict whether the pair matches.

During the training process, for each positive sample pair $(p_i, t_i)$, we randomly construct two negative sample pairs: $(p_i, t_i')$ and $(p_i'', t_i)$. Let $f(p, t) \in [0, 1]$ denote the probability predicted by the Q-Former that the protein-text pair is a match. A corresponding loss function can then be formulated to encourage the model to more accurately distinguish between matching and non-matching samples:

\begin{equation}
\mathcal{L}_{\mathrm{PTM}}=\frac{1}{B}\sum_{i=1}^B[\log f(\boldsymbol{p}_{i^{\prime}},\boldsymbol{t}_i)+\mathrm{log}f(\boldsymbol{p}_i,\boldsymbol{t}_{i^{\prime\prime}})-\log f(\boldsymbol{p}_i,\boldsymbol{t}_i)]
\end{equation}
In this stage, we conducted training using 8 NVIDIA A800 (80GB) GPUs for 30 epochs.

\subsection{End-to-End Fine-tuning}
In this phase, we perform end-to-end fine-tuning of the protein-to-text generation framework. The training data is directly derived from the Swiss-Prot dataset used in the previous stage, without any additional collection or curation of downstream benchmark-specific data. First, the protein sequence is projected into the language model's hidden dimension via a protein token extractor and a learnable linear mapper. Subsequently, this projected representation is appended to the beginning of the tokenized text input, forming the final input sequence passed to the large language model.

Through supervised training on these protein-text pairs, the model learns to generate biologically relevant responses grounded in protein sequence information, thereby enhancing its capacity for various domain-specific tasks.

This setup allows the LLM to directly attend to protein representations during the text generation process without requiring modifications to its internal architecture. By jointly training across the three stages—protein-text alignment, DICP, and supervised fine-tuning—we ultimately obtain a domain-specific large language model. Importantly, no downstream task related data has been training; nevertheless, the model demonstrates strong performance across all evaluated benchmarks. These capabilities emerge naturally from the successful alignment of protein sequences with language representations.

\section{EXPERIMENT AND RESULT}
\subsection{Benchmarks and Metrics}
\textbf{Benchmarks:} We evaluate BioBridge using PFMBench\cite{b30}, a benchmark comprising 16 protein-related tasks across six categories: Annotation, Solubility, Localization, Mutation, Interaction, and Production.

\textbf{Metrics:} We employ three evaluation metrics according to the task type defined in PFMBench: Accuracy (for classification tasks like localization and production), F1 Score (for functional annotation), and Spearman correlation (for regression tasks like solubility, mutation, and interaction).

\begin{table*}[htbp]
    \caption{Comparison of BioBridge and baselines on PFMBench under adapter tuning across core tasks. Qwen means Qwen2.5-7B-Instruct.}
    \label{tab:full_model_performance_adapter}
    \renewcommand{\arraystretch}{1.1}  
    \resizebox{\textwidth}{!}{
        \begin{tabular}{llllllllllllllr} \toprule
        &Dataset         & ESM2   & ProtGPT2 & PGLM    & ProtT5  & DPLM    & SaProt  & ProstT5 & ESM3    & ProTrek & Qwen &\textbf{BioBridge} \\\midrule
      \multirow{2}{*}{\centering Localization }  &DL Bin.       & 0.90619 & 0.90117  & 0.91495 & 0.90736 & 0.93305 & 0.92042 & 0.91657 & 0.90032 & \textbf{0.94336} & 0.56434 &   0.92689 \\
        &DL Multi      & 0.75899 & 0.68442  & 0.72437 & 0.69907 & 0.78029 & 0.69241 & 0.73236 & 0.62051 & 0.80826 &0.10386  &\textbf{0.81516}  \\
        \midrule
        \multirow{3}{*}{\centering Solubility} &DeepSol       & \textbf{0.84494} & 0.78883  & 0.82160 & 0.78741 & 0.82841 & 0.84364 & 0.81937 & 0.78106 & 0.83427 & 0.37655 &0.82882  \\
        &ProtSolM      & \textbf{0.85874} & 0.79735  & 0.84894 & 0.80456 & 0.84847 & 0.85718 & 0.84728 & 0.80773 & 0.83168 &0.51303 &0.76870  \\
        &DeepSoluE     & \textbf{0.77630} & 0.68645  & 0.75549 & 0.72004 & 0.74118 & 0.75492 & 0.74905 & 0.67909 & 0.73090 & 0.52270 &0.62294  \\
        \midrule
         \multirow{4}{*}{\centering Mutation}&FLIP\_AAV     & 0.93848 & 0.33732  & 0.87888 & 0.93825 & \textbf{0.94491} & 0.94822 & 0.93977 & 0.92514 & 0.93999& 0.0000 &0.81355 \\
        &FLIP\_GB1     & 0.95306 & 0.86281  & 0.91945 & 0.95217 & 0.92162 & 0.95133 & \textbf{0.95408} & 0.88144 & 0.94049& 0.0000& 0.75800 \\
        &Stability     & 0.32112 & 0.14803  & \textbf{0.33127} & 0.18638 & 0.29440 & 0.24804 & 0.13032 & 0.15650 & 0.04924& 0.00933& 0.05507  \\
        &Fluo.         & 0.68116 & 0.61042  & 0.66926 & 0.67662 & 0.67930 & \textbf{0.69642} & 0.68020 & 0.66469 & 0.66987 & 0.0227 &0.48597 \\
        \midrule
        \multirow{2}{*}{\centering Interaction}&BindingDB     & 0.13692 & 0.17169  & 0.16884 & 0.19730 & 0.17408 & 0.16557 & 0.16642 & \textbf{0.22519} & 0.19230 &0.05281 &0.17150 \\
        &M. I. Bin.    & 0.71170 & 0.71170  & 0.74513 & 0.72145 & 0.70056 & 0.76323 & 0.72145 & 0.70334 & \textbf{0.80362} &0.34680 &0.76111 \\
        \midrule
         \multirow{1}{*}{\centering Annotation}&EC            & 0.73578 & 0.69687  & 0.74659 & 0.76201 & 0.75521 & 0.75144 & \textbf{0.76829} & 0.64830 & 0.76408 & 0.00272&0.74252 \\
         \midrule
         \multirow{4}{*}{\centering Production}&Thermo.       & 0.95036 & 0.91401  & 0.94224 & 0.92826 & 0.93949 & \textbf{0.96930} & 0.91747 & 0.87837 & 0.93172 &0.60400 &0.88815  \\
        &Anti. Res     & 0.63422 & 0.68437  & 0.67257 & 0.68732 & 0.68732 & 0.65782 & \textbf{0.69027} & 0.58407 & 0.59292 &0.41003 & 0.60465  \\
        &Mat. Pro.     & 0.81189 & 0.76757  & 0.79495 & 0.80072 & 0.80144 & 0.81081 & 0.81622 & 0.77514 & \textbf{0.81477} &0.68793 & 0.80584 \\
        &Clo. CLF      & 0.80586 & 0.77730  & \textbf{0.83638} & 0.78485 & 0.81247 & 0.81206 & 0.79853 & 0.77391 & 0.82612& 0.46760 &0.72532 \\ 
        \bottomrule
        \end{tabular}
    }
\label{tab:result_bio}
\end{table*} 

\begin{table*}[htbp]
    \caption{Comparison of BioBridge and baselines on PFMBench under adapter tuning across core tasks. Qwen means Qwen2.5-7B-Instruct.}
    \label{tab:full_model_performance_adapter}
    \renewcommand{\arraystretch}{1.1}  
    \resizebox{\textwidth}{!}{
        \begin{tabular}{lllllllllllr} \toprule
        &Dataset         & ESM2   & ProtGPT2 & PGLM    & DPLM    & SaProt  & ProstT5 & Qwen &\textbf{BioBridge} \\\midrule
        \multirow{2}{*}{\centering Interaction}&BindingDB     & 0.13692 & 0.17169  & 0.16884  & 0.17408 & 0.16557 & 0.16642 &0.05281 &0.17150 \\
        &M. I. Bin.    & 0.71170 & 0.71170  & 0.74513 & 0.70056 & 0.76323 & 0.72145 & 0.34680 &0.76111 \\
         
        \bottomrule
        \end{tabular}
    }
\label{tab:result_bio1}
\end{table*}

\subsection{Performance Comparison on Protein Property Prediction}
We compared our approach with several methods, including  ESM2\cite{b6}, ESM3\cite{b36}, ProtGPT2\cite{b31}, PGLM\cite{b32}, ProtT5\cite{b33}, DPLM\cite{b17}, ProstT5\cite{b35}, SaProt\cite{b34}, and ProTrek\cite{b15}.

As shown in Table \ref{tab:result_bio}, BioBridge's performance is comparable to foundational protein models on key tasks like annotation (EC), localization (DL Bin./Multi), and interaction (M.I. Bin., BindingDB), with stable scores around 0.74–0.76.  Notably, in the localization task DL Multi and the metal ion binding task M.I. Bin., the model achieved performance improvements of approximately 7\% and 3.5\% respectively on average, demonstrating strong generalization capabilities and domain adaptability.

While slightly weaker than specialized models in solubility, production, and mutation tasks, BioBridge's overall performance is close to, or even partially superior to, foundational models. 
Thanks to continual pre-training and cross-modal alignment, BioBridge shows high versatility and adaptability, effectively replacing the need for specialized protein pre-training.

\begin{table}[htbp]
\small
\caption{General benchmark results.}
\begin{center}

\begin{tabular}{ccccc} \toprule
Benchmark     & Qwen2.5-7B  & ProtT3 & Prot2Chat  & \textbf{BioBridge}   
\\\midrule
agieval       & \textbf{46.03} & 3.63  & 16.12 & 39.82 \\
mmlu          & \textbf{70.41} & 4.29  & 24.48 & 63.30  \\
race          & \textbf{86.51} & 3.85  & 21.58 &  84.00 \\
openbookqa    & 87.80 & 8.75  & 25.60 & \textbf{88.40}  \\
\bottomrule
\end{tabular}
\end{center}
\label{tab:result_general}
\end{table}

\subsection{General Language Capability}

To validate the model's general capabilities, we compare it with representative question-answering protein language models, ProtT3 and Prot2Chat, as well as the general-purpose large language model Qwen2.5-7B-Instruct.

As shown in Table \ref{tab:result_general}, BioBridge exhibits excellent versatility in multi-task scenarios. Its overall performance is on par with, or surpasses on key datasets, the general-purpose large language model Qwen. Conversely, traditional protein language models show weaker cross-task adaptability. The above results confirm that BioBridge possesses stronger cross-task generalization ability and robustness while maintaining its specialization, making it suitable for a wider range of protein-related tasks.

\subsection{Ablation Study}

To evaluate the effectiveness of each component, we designed two ablation experiments comparing our full model against modified variants across protein property prediction tasks.

\begin{itemize}
    \item No Biological Pre-training: We omitted biological pre-training on the LLM but retained the ESM2 encoder and the projector. This assesses if the LLM's intrinsic biological understanding is essential, even with well-aligned protein inputs.
    \item No Protein Encoder/Projector: We removed both the ESM2 encoder and the projector. Raw protein sequences were treated as plain text and fed directly into the LLM for supervised fine-tuning (SFT).
\end{itemize}

\begin{table}[t]
\centering
\small
\caption{Performance Comparison Across Model Variants}
\label{tab:ablation}
\begin{tabular}{lccc}
\toprule
\textbf{Model} & \textbf{DL Bin.} & \textbf{M.I.Bin} & \textbf{Mat.Pro} \\
\midrule
BioBridge & \textbf{0.9269} & \textbf{0.7611} & \textbf{0.8058} \\
w/o pretraining & 0.8958 & 0.6622 & 0.7702 \\
w/o alignment & 0.8113 & 0.3287 & 0.7225 \\
\bottomrule
\end{tabular}
\label{tab:ablation}
\end{table}

\textbf{Analysis:} As shown in the table \ref{tab:ablation}, These ablation experiments assess the criticality of our model's components.  The first setup determines the necessity of the LLM's foundational biological understanding for effective biological reasoning. The second experiment, where raw sequences are used, demonstrates the indispensable contribution of ESM2's specialized protein encoding and the projector's cross-modal alignment, proving these modules are fundamental for the language model to effectively process and reason over complex protein data.



\section{Conclusion}
In this study, we propose BioBridge, a domain-adaptive continual pretraining and semantically-structurally enhanced multi-task protein understanding framework. It introduces biological knowledge through domain-incremental continual pre-training and maps protein sequences into the large language model's semantic space via a protein-text alignment module, thereby enhancing the LLM's ability to comprehend protein structures and functions.

Extensive experiments show that our model performs comparably to specialized protein foundation models across various tasks, yet exhibits stronger generalization and robustness on general datasets, outperforming or matching general large language models like Qwen on multiple tasks. Ablation studies confirm the importance of DICP and cross-modal alignment in this performance boost.

In summary, BioBridge offers a unified and scalable solution, effectively bridging general LLMs with domain-specific protein requirements. This framework holds significant potential for accelerating drug discovery, improving target identification, and gaining deeper insights into protein biology.

\section{Availability of data and materials}
The code and model weights for this study can be found at https://github.com/Yuccaaa/biobridge.

\vspace{12pt}

\end{document}